\pdfoutput=1
\documentclass[11pt,a4paper]{article}
\usepackage{times,latexsym}
\usepackage{url}
\usepackage[T1]{fontenc}
\usepackage{xcolor}
\usepackage{enumitem}
\usepackage{amsmath}
\usepackage{colortbl}
\usepackage[linesnumbered,ruled,vlined]{algorithm2e}
\SetKwComment{Comment}{// }{}
\usepackage[acceptedWithA]{tacl2021v1}

\usepackage{tcolorbox}
\usepackage{xurl}

\newtheorem{definition}{Definition}

\newcommand{\LA}{Link-Append}
\newcommand{\M}{Mention-Link-Append}
\newcommand{\LO}{Link-only}

\usepackage{xspace,mfirstuc,tabulary}

\newif\iftaclinstructions
\taclinstructionsfalse 
\iftaclinstructions
\renewcommand{\confidential}{}
\renewcommand{\anonsubtext}{(No author info supplied here, for consistency with
TACL-submission anonymization requirements)}
\newcommand{\instr}
\fi

\iftaclpubformat 

\else

\fi

\title{Coreference Resolution through a seq2seq Transition-Based System}

\author{
  Bernd Bohnet, Chris Alberti, Michael Collins
  \\ \\
  Google Research
  \\
  \texttt{\{bohnetbd,chrisalberti,mjcollins\}@google.com}
 
}

\date{}

\begin{document}
\maketitle
\begin{abstract}
Most recent coreference resolution systems use search algorithms over possible spans to identify mentions and resolve coreference. We instead present a coreference resolution system that uses a text-to-text (seq2seq) paradigm to predict mentions and links jointly. We implement the coreference system as a transition system and use multilingual T5 as an underlying language model. We obtain state-of-the-art accuracy on the CoNLL-2012 datasets with 83.3 F1-score for English (a 2.3 higher F1-score than previous work \cite{dobrovolskii-2021-word}) using only CoNLL data for training, 68.5 F1-score for Arabic (+4.1 higher than previous work) and 74.3 F1-score for Chinese (+5.3). In addition we use the SemEval-2010 data sets for experiments in the zero-shot setting, a few-shot setting, and supervised setting using all available training data. We get substantially higher zero-shot F1-scores for 3 out of 4 languages than previous approaches and significantly exceed previous supervised state-of-the-art results for all five tested languages. We provide the code and models as open source\footnote{\url{https://github.com/google-research/google-research/tree/master/coref_mt5}}.
\end{abstract}

\section{Introduction}

There has been a great deal of recent research in pretrained language models that employ encoder-decoder or decoder-only architectures (e.g., see see GPT-3, GLAM, Lamda \cite{DBLP:journals/corr/abs-2005-14165, DBLP:journals/corr/abs-2112-06905,DBLP:journals/corr/abs-2201-08239}), and that can generate text using autoregressive or text-to-text (seq2seq) models (e.g., see T5, MT5 \cite{DBLP:journals/corr/abs-1910-10683,xue-etal-2021-mt5}). These models have led to remarkable results on a number of problems. 

\begin{figure}[ht!]
{\small
\noindent
    
\begin{tcolorbox}[colback=blue!1, outer arc=0mm,left=0mm, right=0mm,top=0mm,bottom=0mm] 
{\bf Input:}       
{\it Speaker-A} I still have n't gone to that fresh French restaurant by your house    

{\bf Prediction}: {\sc Shift}: next sentence
\end{tcolorbox}     

\vspace{-0.3cm}
\begin{tcolorbox}[colback=blue!1, outer arc=0mm, left=0mm,right=0mm,top=0mm,bottom=0mm]
{\bf Input}:       
{\it Speaker-A} I$_2$ still have n't gone to that fresh French restaurant by your house {\it Speaker-A} I$_{17}$ 'm like dying to go there

{\bf Prediction}: 
\begin{enumerate}[label=\Alph*,  labelwidth=!, labelindent=0pt, itemsep=0pt, parsep=0pt, topsep=1pt]
\item     I$_{17} \rightarrow$ I$_2$ 
\item     {\sc Shift}: next sentence
\end{enumerate}
\end{tcolorbox}

\vspace{-0.3cm}
\begin{tcolorbox}[colback=blue!1, outer arc=0mm, left=0mm,right=0mm,top=0mm,bottom=0mm]
{\bf Input}:
{\it Speaker-A} \textcolor{blue}{[1 I ]} still have n't gone to that fresh French restaurant by your house {\it Speaker-A} \textcolor{blue}{[1 I ]} 'm like dying to go there {\it Speaker-B} You mean the one right next to the apartment 

{\bf Prediction:} 
\begin{enumerate}[label=\Alph*, labelwidth=!, labelindent=0pt, itemsep=0pt, parsep=0pt, topsep=0pt]
\item You $\rightarrow$ \textcolor{blue}{[1}  \item the apartment  $\rightarrow$ your house  \item the one right next to the apartment $\rightarrow$ that fresh French restaurant by your house 
\item {\sc Shift}: next sentence
\end{enumerate}
\end{tcolorbox} 

\vspace{-0.3cm}
\begin{tcolorbox}[colback=blue!1, outer arc=0mm, left=0mm,right=0mm,top=0mm,bottom=0mm]
{\bf Input}: {\it Speaker-A} \textcolor{blue}{[1 I ]} still have n't gone to \textcolor{orange}{[3 that fresh French restaurant by \textcolor{red}{ [2 your house ]} ]} {\it Speaker-A} \textcolor{blue}{[1 I ]} 'm like dying to go there {\it Speaker-B} \textcolor{blue}{[1 You ]} mean \textcolor{orange}{[3 the one right next to \textcolor{red}{[2 the apartment ]} ] } {\it Speaker-B} yeah yeah yeah

{\bf Prediction}: {\sc Shift}: next sentence
\end{tcolorbox} 
}
 
\caption{ Example of one of our transition-based coreference systems, the {\em \LA} system. The system processes a single sentence at a time, using an input encoding of the prior sentences annotated with coreference clusters, followed by the new sentence. As output, the system makes predictions that link mentions in the new sentence to either previously created coreference clusters (e.g., "You $\rightarrow$ \textcolor{blue}{[1}") or when a new cluster is created, to previous mentions (e.g., "the apartment  $\rightarrow$ your house"). The system predicts "SHIFT" when processing of the sentence is complete. Note in the figure we use the word indices 2 and 17 to distinguish the two incidences of "I" in the text.
    }
    \label{fig:example_introduction}
\end{figure}

Coreference resolution is the task of finding referring expressions in text that point to the same entity in the real world. Coreference resolution is a core task in NLP, relevant to a wide range of applications (e.g., see \newcite{Jurafsky21} Chapter 21 for discussion), but somewhat surprisingly, there has been relatively limited work on coreference resolution using encoder-decoder or decoder-only architectures. 

The state-of-the-art models on coreference problems are based on encoder-only models, such as BERT \cite{devlin-etal-2019-bert} or
SpanBERT \cite{joshi-etal-2020-spanbert}.
All recent state-of-the-art coreference models (see Table \ref{tab:results}) however have the disadvantage of
a) requiring engineering of a specialized search or structured prediction step for coreference resolution, on top of the encoder's output representations; b) often requiring a pipelined approach with intermediate stages of prediction (e.g., mention detection followed by coreference prediction); and c) an inability to leverage more recent work in pretrained seq2seq models.

This paper describes a text-to-text (seq2seq) approach to coreference resolution that can directly leverage modern encoder-decoder or decoder-only models. The method takes as input a sentence at a time, together with prior context, encoded as a string, and makes predictions corresponding to coreference links. The method has the following advantages over previous approaches:

\begin{itemize}
    \item {\bf Simplicity:} We use greedy seq2seq prediction without a separate mention detection step and do not employ a higher order decoder to identify links. 

    \item {\bf Accuracy:} The accuracy of the method exceeds the previous state of the art. 
    \item {\bf Text-to-text (seq2seq) based:} the method can make direct use of modern generation models that employ the generation of text strings as the key primitive.
\end{itemize}

A key question that we address in our work is how to frame coreference resolution as a seq2seq problem. We describe three transition systems, where the seq2seq model takes a single sentence as input, and outputs an action corresponding to a set of coreference links involving that sentence as its output. Figure 1 gives an overview of the highest performing system, “\LA”, which encodes prior coreference decisions in the input to the seq2seq model, and predicts new conference links (either to existing clusters, or creating a new cluster) as its output.
We provide the code and models as open source\footnote{\url{https://github.com/google-research/google-research/tree/master/coref_mt5}}.
Section~\ref{sec:experiments} describes ablations considering other systems, such as a "\LO" system (which does not encode previous coreference decisions in the input), and mention-based (\M), which has a separate mention detection system, in some sense mirroring prior work (see Section ~\ref{sec:multilingual}). 

We describe results on the CoNLL-2012 data set in Section~\ref{sec:experiments}.
In addition,  Section~\ref{sec:multilingual} describes multilingual results, in two settings: first, the setting where we fine-tune on each language of interest; second, zero-shot results, where an MT5 model fine-tuned on English alone is applied to languages other than English. 
Zero-shot experiments show that for most languages, accuracies are higher than recent translation-based approaches and early supervised systems. 


\section{Related Work}
\label{sec:related}

Most similar to our approach is the work of \newcite{webster-curran-2014-limited} which uses a shift-reduce transition-based system for coreference resolution. The transition system uses two data structures, a queue initialized with all mentions and a list. The {\sc Shift} transition moves from the queue a mention to top of the list. The {\sc Reduce} transition merges the top mentions with selected clusters. \newcite{webster-curran-2014-limited} consider the approach to better reflect human cognitive processing, to be simple and to have small memory requirements. \newcite{xia-etal-2020-incremental} use this transition-based system together with a neural approach for mention identification and transition prediction; this neural model \cite{xia-etal-2020-incremental} gives higher accuracy scores (see Table \ref{tab:results}) than \newcite{webster-curran-2014-limited}. 

\newcite{lee-etal-2017-end} focus on predicting mentions and spans using an end-to-end neural model based on LSTMs \cite{HochSchm97}, while \newcite{lee-etal-2018-higher} extends this to a differentiable higher-order model considering directed paths in the antecedent tree. 

Another important method to gain higher accuracy is to use stronger pretrained language models which we follow in this paper as well. A number of recent coreference resolution systems kept the essential architecture fixed while they replace the pretrained models with increasingly stronger models. \newcite{lee-etal-2018-higher} used {\em Elmo} \cite{peters-etal-2018-deep} including feature tuning and show an impressive improvement of 5.1 F1 on the English CoNLL 2012 test set over the baseline score of \newcite{lee-etal-2017-end}. The extension from an end-to-end to the differentiable higher-order inference provides an additional 0.7 F1-score on the test set which leads to a final F1-score of 73.0 for this approach. \newcite{joshi-etal-2019-bert} use the same inference model and explore how to best use and gain another significant improvement of 3.9 points absolute and reach a score of 76.9 F1-score on the test set (see Table \ref{tab:results}). Finally \newcite{joshi-etal-2020-spanbert} use SpanBERT which leads to a even higher accuracy score of 79.6. SpanBERT performs well for coreference resolution due to its span-based pretraining objective.

 \newcite{dobrovolskii-2021-word} considers coreference links between words instead of spans which reduces the complexity to $O(n^2)$ of the coreference models and uses RoBERTa as language model which provides better results than SpanBERT for many tasks.
 
 Similarly, \newcite{kirstain-etal-2021-coreference} reduce the high memory footprint of mention detection by using the start- and end-points of mention spans to identify mentions with a bilinear scoring function. The top $\lambda n$ scored mentions are used to restrict the search space for coreferences prediction using again a bilinear function for scoring. The algorithm has a quadratic complexity since each possible coreference pair has to be scored. 

\newcite{wu-etal-2020-corefqa} cast coreference resolution as question answering and report gains originating from pretraining on Quoref and SQuAD 2.0 of 1 F1-score on the development set. The approach first predicts mentions with a recall-oriented objective, then creates queries for these potential mentions for the cluster prediction. This procedure requires the application of the model for each mention candidate multiple times per document which leads to high execution time.

Our work makes direct use of T5-based models  \cite{DBLP:journals/corr/abs-1910-10683}.
{\sc T5} adopts the idea of treating tasks in Natural Language Processing uniformly as "text-to-text" problems, which means to only have text as input and generate text as output. This idea simplifies and unifies the approach for a large number of tasks by applying the same model, objective, training procedure and decoding process.

\section{Three seq2seq Transition Systems}
\label{sec:systems}

\newcommand{\mi}{{\cal M}_i}
\newcommand{\mm}{{\cal M}}
\newcommand{\mli}{{\cal M}_{\leq i}}

\subsection{The \LA~System}

The \LA~system processes the document a single sentence at a time. At each point the input to the seq2seq model is a text string that encodes the first $i$ sentences together with coreference clusters that have been built up over the first $(i-1)$ sentences. As an example, the input for $i=3$ for the example in Figure~\ref{fig:example_introduction} is the following:

\vspace{2ex}
\noindent
{\bf Input}:
{\it Speaker-A} \textcolor{blue}{[1 I ]} still have n't gone to that fresh French restaurant by your house \# {\it Speaker-A} \textcolor{blue}{[1 I ]} 'm like dying to go there | \# {\it Speaker-B} You mean the one right next to the apartment **

$\;$
\vspace{2ex}

Here the \# symbol is used to delimit sentences, and the start of the focus sentence is marked using the pipe-symbol | and the end of a sentence with two asterisk symbols **.

We have three sentences ($i = 3$). There is a single coreference cluster in the first $i-1 = 2$ sentences, marked using the [1 $\ldots$] bracketings. 

The output from the seq2seq model is also a text string. The text string encodes a sequence of $0$ or more actions, terminated by the SHIFT token. Each action links some mention (a span) in the $i$th sentence to some mention in the previous context (often in the first $i-1$ sentences, but sometimes also in the $i$th sentence). An example prediction given the above input is the following:

\vspace{2ex}
\noindent
{\bf Prediction} You $\rightarrow$ \textcolor{blue}{[1} ; the apartment  $\rightarrow$ your house; the one right next to the apartment $\rightarrow$ that fresh French restaurant by your house ;    {\sc Shift}\\

\noindent
More precisely, the first action would actually be "You \#\# mean the one $\rightarrow$ [1" where the substring "mean the one" is the 3-gram in the original text immediately after the mention "You". The 3-gram helps to disambiguate the mention fully, in the case where the same string might appear multiple times in the sentence of interest. For brevity we omit these 3-grams in the following discussion, but they are used throughout the models output to specify mentions.\footnote{Note that no explicit constraints are placed on the model's output, so there is the potential for the model to generate mention references that do not correspond to substrings within the input; however this happens very rarely in practice, see section~\ref{sec:error-analysis} for discussion. There is also the potential for the 3-gram to be insufficient context to disambiguate the exact location of a mention; again, this happens rarely, see section~\ref{sec:error-analysis}.}

$\;$
\vspace{2ex}

\noindent
In this case there are three actions, separated by the ";" symbol, followed by the terminating {\sc SHIFT} action. The first action is

\vspace{2ex}
\noindent
You $\rightarrow$ \textcolor{blue}{[1}

\vspace{2ex}
\noindent
This is an {\bf append} action: specifically, it appends the mention "You" in the 3rd sentence to the existing coreference cluster labeled [1 $\ldots$]. The second action is

\vspace{2ex}
\noindent
the apartment  $\rightarrow$ your house

\vspace{2ex}
\noindent
This is a {\bf link} action. It links the mention "the apartment" in the 3rd sentence to "your house" in the previous context. Similarly the third action,

\vspace{2ex}
\noindent
the one right next to the apartment $\rightarrow$ that fresh French restaurant by your house

\vspace{2ex}
\noindent
is a also a link action, in this case linking the mention "the one right next to the apartment" to a previous mention in the discourse.

The sequence of actions is terminated by the {\sc SHIFT} symbol. At this point the $i$th sentence has been processed, and the model moves to the next step where the $(i+1)$th sentence will be processed. Assuming the next sentence is "{\it Speaker-B} yeah yeah yeah", the input at the $(i+1)$th step will be

\vspace{2ex}
\noindent
{\bf Input}:
{\it Speaker-A} \textcolor{blue}{[1 I ]} still have n't gone to \textcolor{orange}{[3 that fresh French restaurant by \textcolor{red}{ [2 your house ]} ]} \# {\it Speaker-A} \textcolor{blue}{[1 I ]} 'm like dying to go there \# {\it Speaker-B} \textcolor{blue}{[1 You ]} mean \textcolor{orange}{[3 the one right next to \textcolor{red}{[2 the apartment ]} ] } | \# {\it Speaker-B} yeah yeah yeah

\vspace{2ex}
\noindent
Note that the three actions in the previous prediction have been reflected in the new input, which now includes three coreference clusters, labeled [1 $\ldots$], [2 $\ldots$] and [3 $\ldots$].

In summary, the method processes a sentence at a time, and uses {\bf append} and {\bf link} actions to build up links between mentions in the current sentence under focus and previous mentions in the discourse.

A critical question is how to map training data examples (which contain coreference clusters for entire documents) to sequences of actions for each sentence. Clearly there is some redundancy in the system, in that in many cases either link or append actions could be used to build up the same set of coreference clusters. We use the following method for creation of training examples:

\begin{itemize}
    \item Process mentions in the order in which they appear in the sentence. Specifically, mentions are processed in order of their end-point (earlier end-points are earlier in the ordering). Ties are broken by their start-point (later start-points are earlier in the ordering). It can be seen that the order in the previous example, {\em You}, {\em the apartment}, {\em the one right next to the apartment}, follows this procedure.
    
    \item For each mention, if there is another mention in the same coreference cluster earlier in the document, either:
    
    \begin{enumerate}
        \item Create an {\em append} action if there are at least two members of the cluster in the previous $i-1$ sentences.
        \item Otherwise create a {\em link} action to the most recent member of the coreference cluster (this may be either in the first $i-1$ sentences, or in the $i$th sentence).
    \end{enumerate}
    
\end{itemize}

The basic idea then will be to always use append actions where possible, but to use link actions where a suitable append action is not available. 

\subsection{The \LO~System}

The \LO~system is a simple variant of the \LA~system. There are two changes: First, the only actions in the \LO~system are link, and SHIFT, as described in the previous section. Second, when encoding the input in the \LO~system, the first $i$ sentences are taken again with the \# separator, but no information about coreference clusters over the first $i-1$ sentences is included.

The \LO~system can therefore be viewed as a simplification of the \LA~system. We will compare the two systems in experiments, in general seeing that the \LA~system gives significant improvements in performance.

\subsection{The \M~System}

The \M~ system is a modification of the \LA~system, which includes an additional class of actions, the {\em mention} actions. A mention action selects a single sub-string from the sentence under focus, and creates a singleton coreference cluster. The algorithm that creates training examples is modified to have an additional step for the creation of {\em mention} actions, as follows: 

\begin{itemize}
    \item Process mentions in the order in which they appear in the sentence. 
    
    \item For each mention, if it is the first mention in a coreference structure, introduce a {\em mention} action for that mention.
    
    \item For each mention, if there is another mention in the same coreference cluster earlier in the document, either:
    
    \begin{enumerate}
        \item Create an {\em append} action if there is at least two members of the cluster in the previous $i-1$ sentences.
        \item Otherwise create a {\em link} action to the most recent member of the coreference cluster (this may be either in the first $i-1$ sentences, or in the $i$th sentence).
    \end{enumerate}
    
\end{itemize}
Note that the \M~system can create singleton coreference structures, unlike the {\sc Link-Append} or \LO~systems. This is its primary motivation.

\subsection{A Formal Description}

We now give a formal definition of the three systems. This section can be safely skipped on a first reading of the paper.

\subsubsection{Initial Definitions, and Problem Statement}

We introduce some key initial definitions---of {\em documents}, {\em potential mentions}, and {\em clusterings}---before giving a {\em problem statement}:

\begin{definition}[Documents]
A document is a pair $(w_1 \ldots w_n, s_1 \ldots s_m)$, where $w_i$ is the $i$th word in the document, and $s_1, s_2, \ldots, s_m$ is a sequence of integers specifying a segmentation of $w_1 \ldots w_n$ into $m$ sentences. Each $s_i$ is the endpoint for sentence $i$ in the document. Hence $1 \leq s_1 < s_2 \ldots s_{m-1} < s_m$, and $s_m = n$. The $i$th sentence spans words $(s_{i-1}+1) \ldots s_i$ inclusive (where for convenience we define $s_0 = 0$).
\end{definition}

\begin{definition}[Potential Mentions]
Assume an input document $(w_1 \ldots w_n, s_1 \ldots s_m)$.
For each $i \in 1 \ldots m$ we define ${\cal M}_i$ to be the set of {\em potential mentions} in the $i$th sentence; specifically,
\[
{\cal M}_i = \{ (a, b): s_{i-1} < a \leq b \leq s_i \}
\]
Hence each member of ${\cal M}_i$ is a pair $(a, b)$ specifying a subspan of the $i$th sentence. We define
\[
{\cal M} = \cup_{i=1}^m {\cal M}_i,
\;\;\;
{\cal M}_{\leq i} = \cup_{j=1}^i {\cal M}_j
\]
hence ${\cal M}$ is the set of all potential mentions in the document, and $\mli$ is the set of potential mentions in sentences $1 \ldots i$.
\end{definition}

\begin{definition}[Clusterings]
A clustering $K$ is a sequence of sets $K_1, K_2, \ldots K_{|K|}$, where each $K_i \subseteq \mm$, and for any $i, j$ such that $i \neq j$, we have $K_i \cap K_j = \emptyset$. We in addition assume that for all $i$, $|K_i| \geq 2$ (although see Section~\ref{sec:three-systems} for discussion of the case where $|K_i| \geq 1$). We define ${\cal K}$ to be the set of all possible clusterings. 
\end{definition}

\begin{definition}[Problem Statement] 
The coreference problem is to take a document $x$ as input, and to predict a clustering $K$ as the output. We assume a training set of $N$ examples, $\{(x^{(i)}, K^{(i)})\}_{i=1}^N$, consisting of documents paired with clusterings.
\end{definition}

\subsection{The Three Transition Systems}
\label{sec:three-systems}

The transition systems considered in this paper take a document $x$ as input, and produce a coreference clustering $K$ as the output. We  assume a definition of transition systems that is closely related to work on deterministic dependency parsing \cite{nivre-2003-efficient,nivre-2008-algorithms}, and which is very similar to the conventional definition of deterministic finite-state machines. Specifically, a transition system consists of: 1) A set of states ${\cal C}$. 2) An initial state $c_0 \in {\cal C}$. 3) A  set of actions ${\cal A}$. 4) A transition function $\delta: C \times {\cal A} \rightarrow C$. This will usually be a partial function: that is for a particular state $c$, there will be some actions $a$ such that $\delta(c, a)$ is undefined. For convenience, for any state $c$ we define ${\cal A}(c) \subseteq {\cal A}$ to be the set of actions such that for all $a \in {\cal A}(c)$, $\delta(c, a)$ is defined. 5) A set of final states ${\cal F} \subseteq {\cal C}$.

A path is then a sequence $c_0, a_0, c_1, a_1, \ldots c_N$ where for $i = 1 \ldots N$, $c_{i+1} = \delta(c_{i}, a_i)$, and where $c_N \in {\cal F}$.

All transition systems in this paper use the following definition of states:
\begin{definition}[States]
\label{defn:states}
A state is a pair $(i, K)$ such that $1 \leq i \leq (m+1)$ and $K \in {\cal K}$ is a clustering such that for $k \in 1 \ldots |K|$, for $j \in (i+1) \ldots m$, $K_k \cap M_j = \emptyset$ (i.e,. $K$ is a clustering over the mentions in the first $i$ sentences). In addition we define the following:

\begin{itemize}
    \item  ${\cal C}$ is the set of all possible states.
    \item $c_0 = (1, \epsilon)$ is the initial state, where $\epsilon$ is the empty sequence.
    \item ${\cal F} = \{ (i, K): (i, k) \in {\cal C}, i = (m+1)\}$ is the set of final states.
\end{itemize}
\end{definition}
Intuitively, the state $(i, K)$ keeps track of which sentence is being worked on, through the index $i$, and also keeps track of a clustering of the partial mentions up to and including sentence $i$. 

We now describe the actions used by the various transition systems. The actions will either augment the clustering $K$, or increment the index $i$. The actions fall into four classes---{\em link actions}, {\em append actions}, {\em mention actions}, and the {\em shift action}---defined as follows:

\paragraph{Link actions.}
Given a state $(i, K)$, we define the set of possible link actions as
\[
L(i, K) = 
\{m \rightarrow m' : m \in {\cal M}_i, m' \in \mli\}
\]
A link action $(m \rightarrow m')$ augments $K$ by adding a link between mentions $m$ and $m'$. We define $K \oplus (m \rightarrow m')$ to be the result of adding link $m \rightarrow m'$ to clustering $K$.\footnote{Specifically, the addition of the link $m \rightarrow m'$ can either: 1) create a new cluster within $K$, if neither $m$ or $m'$ are in an existing cluster within $K$; 2) add $m$ to an existing cluster within $K$, if $m'$ is already in some cluster in $K$, and $m$ is not in an existing clustering; 3) add $m'$ to an existing cluster within $K$, if $m$ is already in some cluster in $K$, and $m'$ is not in an existing clustering; 4) merge two clusters, if $m$ and $m'$ are both in clusters within $K$, and the two clusters are different; 5) leave $K$ unchanged, if $m$ and $m'$ are both within the same existing cluster within $K$. In practice cases (2), (3), (4) and (5) are never seen in oracle sequences of actions, but for completeness we include them.}
We can then define the transition function associated with a link action:
\[
\delta((i,K), m \rightarrow m') = (i, K \oplus (m \rightarrow m'))
\]

\paragraph{Append actions.}
Given a state $(i, K)$, we define the set of possible append actions as
\[
\hbox{App}(i, K) = 
\{m \rightarrow k : m \in {\cal M}_i, k \in \{1 \ldots |K|\}\}
\]
An append action $(m \rightarrow k)$ augments $K$ by adding mention $m$ to the cluster $K_k$ withing the sequence $K$. We define $K \oplus (m \rightarrow k)$ to be the result of this action (thereby overloading the $\oplus$ operator); the transition function associated with an append action is then
\[
\delta((i,K), m \rightarrow k) = (i, K \oplus (m \rightarrow k))
\]

\paragraph{Mention actions.}
Given a state $(i, K)$, we define the set of possible mention actions as
\[
\hbox{Mention}(i, K) = 
\{\hbox{Add}(m) : m \in {\cal M}_i\}\}
\]
A mention action $\hbox{Add}(m)$ augments $K$ by either creating a new singleton cluster containing $m$ alone, assuming that $m$ does not currently appear in $K$; otherwise it leaves $K$ unchanged. We define $K \oplus \hbox{Add}(m)$  to be the result of this action, and $\delta((i,K), \hbox{Add}(m)) = (i, K \oplus \hbox{Add}(m))$.

\paragraph{The SHIFT action.}
The final action in the system is the SHIFT action. This can be applied in any state, and simply advances the index $i$, leaving the clustering $K$ unchanged:
\[
\delta((i,K), \hbox{SHIFT}) = ( (i+1), K)
\]

We are now in a position to define the transition systems:

\begin{definition}[The three transition systems]
The link-append transition system is defined as follows:
\begin{itemize}
    \item ${\cal C}$, $c_0$, and ${\cal F}$ are as defined in definition~\ref{defn:states}.
    \item For any state $(i, K)$, the set of possible actions is ${\cal A}(i, K) = L(i, K) \cup \hbox{App}(i, K) \cup \{\hbox{SHIFT}\}$. The full set of actions is ${\cal A} = \cup_{(i, K) \in {\cal C}} {\cal A}(i, K)$
    \item The transition function $\delta$ is as defined above.
\end{itemize}

The \LO~system is identical to the above, but with ${\cal A}(i, K) = L(i, K) \cup  \{\hbox{SHIFT}\}$. The \M~system is identical to the above, but with ${\cal A}(i, K) = L(i, K) \cup \hbox{App}(i, K) \cup \hbox{Mention}(i, k) \cup \{\hbox{SHIFT}\}$.

\end{definition}

All that remains in defining the seq2seq method for each transition system is to: a) define an encoding of the state $(i, K)$ as a string input to the seq2seq model; b) define an encoding of each type of action, and of a sequence of actions corresponding to single sentence; c) defining a mapping from a training example consisting of an $(x, K)$ pair to a sequence of input-output texts corresponding to training examples. 

\section{Experimental Setup}
\label{sec:experiments}

\definecolor{Gray}{gray}{0.9}

\begin{table}
\centering
\small
\setlength{\tabcolsep}{3.0pt}
\begin{tabular}{l|rr|rr|rr}
& \multicolumn{2}{c|}{Training} & \multicolumn{2}{c|}{Development} & \multicolumn{2}{c}{Test} \\
Language & docs & tokens & docs & tokens & docs & tokens  \\ \hline
\rowcolor{Gray}
\multicolumn{7}{c}{\bf OntoNotes / CoNLL-2012 datasets} \\ 
English & 1940 & 1.3M & 343 & 160k & 348 & 170k   \\
Chinese & 1729 & 750k & 254 & 110k & 218 & 90k\\
Arabic & 359 & 300k & 44 & 30k & 44 & 30k \\ \hline
\rowcolor{Gray}
\multicolumn{7}{c}{\bf SemEval 2010 data} \\
Catalan  & 829  & 253k   & 142& 42k & 167  & 49k\\
Dutch    & 145  & 46k    & 23 & 9k &72 & 48k \\
German   & 900  & 331k   & 199   & 73k &136 & 50k \\
Italian  & 80  &   81k   & 18    & 16k& 46 & 41k \\
Spanish  & 875 & 284k    &140    & 44k& 168 & 51k
\end{tabular}

\caption{\label{tab:datasets} Sizes of the SemEval Shared Task data sets and OntoNotes (CoNLL-2012). }

\end{table}


\begin{table*}
\centering
\small
\renewcommand{\arraystretch}{1}
\setlength{\tabcolsep}{3.5pt}
\begin{tabular}{l|l|l|ccc|ccc|ccc|c}

 & & & \multicolumn{3}{c|}{MUC} & \multicolumn{3}{c|}{B$^{3}$} & \multicolumn{3}{c|}{CEAF$_{\Phi_4}$} & Avg. \\ 
  & LM  & Decoder &P & R & F1 & P & R & F1 & P & R & F1 & F1 \\ \hline
\rowcolor{Gray}
\multicolumn{13}{c}{\bf English} \\
\newcite{lee-etal-2017-end}       & - & neural e2e & 78.4 & 73.4 & 75.8& 68.6 & 61.8 & 65.0 & 62.7 & 59.0 & 60.8 & 67.2 \\
\newcite{lee-etal-2018-higher}    &  Elmo & c2f & 81.4 & 79.5 & 80.4& 72.2 & 69.5 & 70.8 & 68.2 & 67.1 & 67.6 & 73.0 \\
\newcite{joshi-etal-2019-bert}    & BERT & c2f & 84.7 & 82.4 & 83.5 & 76.5 & 74.0 & 75.3 & 74.1 & 69.8 & 71.9 & 76.9\\
\newcite{yu-etal-2020-cluster} & BERT & Ranking & 82.7 & 83.3 & 83.0 & 73.8 & 75.6 &  74.7 &  72.2 & 71.0 & 71.6 & 76.4 \\
\newcite{joshi-etal-2020-spanbert}& SpanBERT & c2f & 85.8 & 84.8 & 85.3 & 78.3 & 77.9 & 78.1 & 76.4 & 74.2 & 75.3 & 79.6 \\
\newcite{xia-etal-2020-incremental}& SpanBERT & transitions & 85.7 & 84.8 & 85.3 & 78.1 & 77.5 & 77.8 & 76.3 & 74.1 & 75.2 & 79.4 \\
\newcite{wu-etal-2020-corefqa}      & SpanBERT & QA & 88.6 & 87.4 & 88.0 & 82.4 & 82.0 & 82.2 & 79.9 & 78.3 & 79.1 & {\em 83.1}$^*$\\
\newcite {xu-choi-2020-revealing} & SpanBERT & hoi & 85.9 &  85.5 & 85.7 & 79.0 & 78.9 & 79.0 & 76.7 & 75.2 & 75.9 & 80.2 \\
\newcite{kirstain-etal-2021-coreference} &LongFormer & bilinear & 86.5 & 85.1 & 85.8 & 80.3 & 77.9 & 79.1 & 76.8 & 75.4&  76.1 & 80.3 \\
\newcite{dobrovolskii-2021-word}  & RoBERTa & c2f & 84.9 & 87.9 &86.3 & 77.4 & 82.6 & 79.9 & 76.1 & 77.1 & 76.6 & 81.0 \\
 
 \LA              & mT5 & transition & 87.4 & 88.3 & 87.8 & 81.8 & 83.4 & 82.6  & 79.1 & 79.9 & 79.5 & \bf 83.3 \\
\hline

\rowcolor{Gray}
\multicolumn{13}{c}{\bf Arabic} \\ 
\newcite{aloraini-etal-2020-neural} & AraBERT & c2f & 63.2 & 70.9 & 66.8 & 57.1 & 66.3 & 61.3 & 61.6 & 65.5 & 63.5 & 63.9 \\ \newcite{min-2021-exploring} &GigaBERT & c2f  &73.6& 61.8& 67.2& 70.7& 55.9& 62.5& 66.1& 62.0& 64.0& 64.6 \\
Link-Append & mT5 & transition & 71.0 & 70.9 & 70.9 & 66.5 & 66.7 & 66.6  & 68.3 & 68.6 & 68.4 & \bf 68.7	\\
\hline

\rowcolor{Gray}
\multicolumn{13}{c}{\bf Chinese} \\ 
\newcite{xia-van-durme-2021-moving} &  XLM-R & transition & - & - & - & - & - & - & - & - & - & 69.0 \\
Link-Append & mT5 & transition & 81.5 & 76.8 & 79.1 & 76.1 & 69.9 & 72.9  & 74.1 & 67.9 & 70.9 & \bf  74.3 \\ 
\end{tabular}

\caption{English, Arabic and Chinese  test set results and comparison with previous work on the CoNLL-2012 Shared Task test data set. The average F1 score of MUC, B$^3$ and CEAF$_{\Phi_4}$ is the main evaluation criterion. $^*$\newcite{wu-etal-2020-corefqa} use additional training data. }
\label{tab:results}

\end{table*}

We train an mT5 model to predict from an input a target text. We use the provided training, development and test splits as described in section \ref{sec:data}. For the preparation of the input text, we follow previous work and include the speaker in the input text before each sentence \cite{wu-etal-2020-corefqa} as well as the text genre at the document start if this information is available in the corpus. We apply as described in Section \ref{sec:systems} the corresponding transitions as an oracle to obtain the input and target texts. We shorten the text at the front if the input text is larger as the sentence piece token input size of the language model and add further context beyond the sentence $i$ when the input space is not filled up (note that as described in section~\ref{sec:systems}, we use the pipe symbol | to mark the start of the focus sentence and the end with two asterisk symbols **).

\subsection{Data}
\label{sec:data}

We use the English coreference resolution dataset from the CoNLL-2012 Shared Task \cite{pradhan-etal-2012-conll} and SemEval-2010 Shared Task set \cite{recasens-etal-2010-semeval} for multilingual coreference resolution experiments.
The statistics on the dataset sizes are summarized in Table \ref{tab:datasets}. The table shows that the English CoNLL-2012 Shared Task is substantially larger than any of the other data sets.

\subsection{Experiments}
\label{experiments}

\paragraph{Setup for English.}
For our experiments, we use mT5 and initialize our model with either the {\tt xl} or {\tt xxl} checkpoints\footnote{\url{https://github.com/google-research/multilingual-t5}}. For fine-tuning, we use the hyperparameters suggest by \newcite{DBLP:journals/corr/abs-1910-10683}: a batch-size of 128 sequences and a constant learning rate of 0.001. We use micro-batches of 8 to reduce the memory requirements. We save checkpoints every 2k steps. From these models, we select the model with the best development results. We train for 100k steps. We use inputs with 2048 sentence piece tokens and 384 output tokens for training. The training of the {\tt xxl}-model takes about 2 days on 128 TPUs-v4. 
On the development set, inference takes about 30 minutes on 8 TPUs.   

\paragraph{Setup for other languages.} We used the English model in this work to continue training with the above settings on other languages than English (Arabic, Chinese and the SemEval-2010 datasets). 
For few-shot learning, we use the first 10 documents for each language and we train only for 200 steps since the evaluation then shows 100\% fit to the training set.

All our models have been tested with 3k sentence piece tokens input length. In Section \ref{sec:multilingual}, we present our work on multilingual coreference resolution and Section \ref{sec:discussion} discusses the results for all languages.


\begin{table}[ht!]
\small
\centering
\setlength{\tabcolsep}{1pt}
\begin{tabular}{l|cc|c|c}
 & \multicolumn{2}{c|}{Sing.}    & \# training    & Avg. \\
  Systems                              & P & E                  &  docs./method         & \multicolumn{1}{c}{F1} \\ \hline
\rowcolor{Gray}
\multicolumn{5}{c}{\bf Catalan} \\

\newcite{attardi-etal-2010-tanl}     &Y& Y & all & 48.2 \\
\M                &Y& Y & all & \bf 83.5 \\ \hline
\newcite{xia-van-durme-2021-moving}  &N& Y & all &51.0 \\ 
\M                &N& Y & all &\bf  59.2 \\
\hline
\newcite{bitew-etal-2021-lazy}       &N& N & $\emptyset$/Translation & \bf 48.0 \\
\LA                        &N& N  & $\emptyset$/Zero-shot & 47.7 \\ \hline
\LA                        &N& N & 10/Few-shot & 68.9\\ 
\hline
\rowcolor{Gray}
\multicolumn{5}{c}{\bf Dutch} \\
\newcite{kobdani-schutze-2010-sucre} &Y& Y & all & 19.1 \\ 
\M                     &Y& Y & all & \bf 66.6 \\\hline
\newcite{xia-van-durme-2021-moving}  &N& Y & all & 55.4 \\ 
\M                     &N& Y & all & \bf 59.9\\ \hline
\newcite{bitew-etal-2021-lazy}       &N& N & $\emptyset$/Translation & 37.5 \\
\LA                        &N& N & $\emptyset$/Zero-shot & \bf 57.6 \\ \hline
\LA                        &N& N & 10/Few-shot & 65.7\\ 

\hline
\rowcolor{Gray}
\multicolumn{5}{c}{\bf German} \\

\newcite{kobdani-schutze-2010-sucre} &Y& Y & all & 59.8 \\
\M                     &Y& Y & all & \bf 86.4 \\ \hline
\newcite{roesiger-kuhn-2016-ims}     &N& N & all & 48.6 \\
\newcite{schroder-etal-2021-neural}  &N& N & all & 74.5 \\ 
\LA                       &N& N & all & \bf 77.8 \\ 
\hline
\LA                       &N& N & $\emptyset$/Zero-shot& 55.0 \\ \hline  
\LA                       &N& N & 10/Few-shot & 69.8 \\ 
\hline
\rowcolor{Gray}
\multicolumn{5}{c}{\bf Italian} \\
\newcite{kobdani-schutze-2010-sucre} &Y& Y & all & 60.7 \\ 
\M                     &Y& Y & all & \bf 65.9 \\ \hline 
\M                     &N& N & all & 59.4 \\ 
\hline
\newcite{bitew-etal-2021-lazy}      &N& N & $\emptyset$/Translation & 36.2 \\ 
\LA                                 &N& N & $\emptyset$/Zero-shot & \bf 39.4\\ \hline
\LA                                 &N& N & 10/Few-shot & 61.2\\

\hline
\rowcolor{Gray}
\multicolumn{5}{c}{\bf Spanish} \\
\newcite{attardi-etal-2010-tanl}    &Y& Y & all & 49.0 \\
\M                &Y& Y & all &  \bf 83.9 \\ \hline
\newcite{xia-van-durme-2021-moving} &N& Y & all & 51.3 \\ 
\M                     &N& Y & all & \bf 59.3 \\ \hline
\LA                       &N& N & all & 83.1\\ \hline 
\newcite{bitew-etal-2021-lazy}      &N& N &$\emptyset$/Translation & 46.1 \\
\LA                       &N& N & $\emptyset$/Zero-shot & \bf 49.4 \\ \hline
\LA                       &N& N & 10/Few-shot & 72.5 \\

\end{tabular}
\caption{Test set results for SemEval-2010 datasets.  The {\em Sing.} column shows whether, the singletons are included (Y) or removed (N) in the {\bf P}rediction and the {\bf E}valuation set. The last column shows average F1 score of MUC, B$^3$ and CEAF$_{\Phi_4}$. 
\label{tab:x-shot}
}
\end{table}

\section{Multilingual Coreference Resolution Results}
\label{sec:multilingual}

The SemEval-2010 datasets \cite{recasens-etal-2010-semeval} include six languages and is therefore a good test bed for multilingual coreference resolution. We excluded English as the data overlaps with our training data. 
The experimental evaluation changed in recent publication by reporting F1-scores as an average of MUC, B$^3$ and CEAF$_{\Phi_4}$ following the CoNLL-2012 evaluation schema. We follow this schema  in this paper as well. 

Another important difference between the SemEval-2010 and the CoNLL-2012 datasets is the annotation of singletons (mentions without antecedents) in the SemEval datasets. 
Most recent systems predict only coreference chains. This has lead also to different evaluation methods for the SemEval-2010 datasets. The first method keeps the singletons for the evaluation purposes (e.g. \cite{xia-van-durme-2021-moving}) and the second excludes the singletons from evaluation set (e.g. \cite{roesiger-kuhn-2016-ims,schroder-etal-2021-neural,bitew-etal-2021-lazy}). The exclusion of singletons seems better suited to compare recent systems but makes direct comparison with previous work difficult. In our results overview (Table \ref{tab:x-shot}), we report in the column {\em sing.} whether singletons are included (Y) or excluded (N), in the P-column for prediction and in the E-column for evaluation.

\subsection{Zero-Shot and Few-Shot}

Since mT5 is pretrained on 100+ languages \cite{xue-etal-2021-mt5}, we evaluate Zero-Shot transfer ability from English to other languages. We apply our system trained on the English CoNLL-2012 Shared Task dataset to the non-English SemEval-2010 test sets. Table \ref{tab:x-shot} shows evaluation scores for our transition-based systems and reference systems. 
We use for training the same setting as the reference systems  \cite{kobdani-schutze-2010-sucre,roesiger-kuhn-2016-ims,schroder-etal-2021-neural}. In the Zero-shot experiments, the transition-based systems are trained only on English CoNLL-2012 datasets and applied without modification to the multilingual SemEval-2010 test sets. 

\newcite{bitew-etal-2021-lazy} use machine translation for the coreferences prediction of the SemEval-2010 datasets. 
The authors found they got the best accuracy when they first translated the test sets to English, then predicted the English coreferences with the system of \newcite{joshi-etal-2020-spanbert} and finally projected back the predictions. They apply this method to four out of the six languages for the SemEval-2010 datasets. We included in Table \ref{tab:x-shot} their results as a comparison to our Zero-Shot results. The two methods are directly comparable as they do not use the target language annotations for training. Our Zero-Shot F1-scores are substantially higher compared with the machine translation approach for Dutch, Italian, Spanish and a bit lower for Catalan, cf. Table \ref{tab:x-shot}. 

\newcite{xia-van-durme-2021-moving} explored for a large number of settings few-shot learning using the continued training approach. We use the same approach with a single setting that uses the first 10 documents for each language. For details about the experimental setup see Section \ref{experiments}. Table \ref{tab:x-shot} presents the results for the Link-Append system. 
This shows that already with a few additional training documents a high accuracy can be reached. This could be useful either to adapt to a specific coreference annotations schema or to specific language (see examples in Figure 2 and 3). 

\subsection{Supervised}

We also carried out experiments in a fully supervised setup in which we use all available training data of the SemEval-2010 Shared Task. We adopted the method of continued training of \newcite{xia-van-durme-2021-moving}. In our experiments, we start from our finetuned English model and continue training on the SemEval-2010 datasets and the Arabic OntoNotes dataset for the later we use data and splits of the CoNLL-2012 Shared Task.  

To verify the finding of \newcite{xia-van-durme-2021-moving}, we compared the results when we continue the training from a finetuned model and from the initial mT5 model. We conducted this exploratory experiments using 1k training steps for the German dataset. The results are in favor of the experiment with continued training using an already fine-tuned model with a score of 84.5 F1 vs 81.0 F1 for fresh mT5 model. This model also achieves 77.3 F1 when evaluated without singletons (cf. Table \ref {tab:x-shot}), surpassing previous SotA of 74.5 F1 \cite{schroder-etal-2021-neural}. We did not explore training longer due to computational cost of training from a fresh mT5 model to reach a potentially better performance. We adopted the approach for all datasets of the SemEval-2010 Shared Task as this model provides competitive coreference models with low training cost. 

 Table \ref {tab:x-shot} includes the accuracy scores for the cluster/mentions-based transition systems which reaches SotA for all languages when the prediction and evaluation includes the singletons (P=Y, E=Y). In order to compare the results with \newcite{xia-van-durme-2021-moving}, we removed from the results for the cluster/mentions-based transition system the singletons in the prediction but still include them in the evaluation (P=N, E=Y).

Table \ref{tab:results} compares the results for Arabic and Chinese of our model with the recent work. The Link-Append system is 4.1 points better than \cite{min-2021-exploring} and 5.3 points better than \cite{xia-van-durme-2021-moving} which presents previous SotA for Arabic and Chinese, respectively.

\section{Discussion}
\label{sec:discussion}

In this section, we analyse performance factors with an ablation study, analyse errors, and reflect on design choices that are important for the model's performance. Table \ref{tab:results} shows the results for our systems on the English, Arabic and Chinese CoNLL-2012 Shared Task and compares with previous work.
\begin{table}[ht]
\centering
\small
\renewcommand{\arraystretch}{1}
\setlength{\tabcolsep}{3.5pt}
\begin{tabular}{l|l|l}
System     & Ablation & F1 \\  \hline
\LA        & 100k steps/3k pieces  &  83.2\\
\LA        & 2k sentence pieces &  83.1\\
\LA        & 50k steps  &  82.9\\
\LA        & no context beyond $i$ &  82.8\\
\LA        & xxl-T5.1.1            &  82.7\\
\LA        & xl-mT5     &  78.0\\ 
\hline
\M        & 3k pieces &   82.6\\ 
\M        & 2k pieces  &  82.2\\ \hline
\LO       & link transitions only  & 81.4 \\ 
\end{tabular}
\caption{{\bf Development set} results for an ablation study using English CoNLL-2012 data sets and reporting Avg. F1-scores. The models have been trained with 100k training steps and tested with 2k sentence pieces filling up remaining space in the input beyond the focus sentence $i$ with further sentences of the document as context. In inference mode, the model uses input length of 3k sentences pieces if not stated otherwise.
\label{tab:ablation}
}
\end{table}

\subsection{Ablation Study}

Our best-performing transition system is Link-Append, which predicts links and clusters iteratively for sentences of a document without predicting mentions before-hand. Table \ref{tab:ablation} shows an ablation study. The results at the top of the table show the development set results for the Link-Append system when, with each {\sc Shift}, the already identified coreference clusters are annotated in the input. This information is then available in the next step and the clusters can be extended by the {\sc Append} transition.

The models are trained with an input size of 2048 tokens using mT5. We use a larger input size of 3000 (3k) tokens for decoding to accommodate for long documents and very long distances between mentions of coreferences. When we use 2k sentence pieces, the accuracy is 83.1 instead of 83.2 averaged F1-score on the development set using the model trained for 100k steps. 

At the bottom of Table \ref{tab:ablation}, the performance of a system is shown that does not annotate the identified clusters in the input. In this system the Append transition cannot be applied and hence only the Link and Shift transition are used. The accuracy of this system is substantially lower, by 1.8 F1-score. 

We observe drops in accuracy when we do not use context beyond the sentence $i$ or when we train for only 50k steps. We observe 0.5 lower F1-score, when we use xxl-T5.1.1\footnote{The xxl-T5.1.1 model refers to a model provided by \newcite{xue-etal-2021-mt5} trained for 1.1 million steps on English data.} instead of the xxl-mT5 model. An analysis shows that the English OntoNotes corpus contains some non-English text, speaker names and special symbols. For instance, there are Arabic names that are mapped to OOV, but also the curly brackets $\{ \}$. There are also other cases where T5 translated non-English words to English (e.g. German 'nicht' to 'not'). 
 
With the \M~ system, we introduced a system that is capable of introducing mentions, which is useful for data sets that include single mentions, such as the SemEval-2010 data set. This transition system has an 82.6 F1-score on the development set with an input context of 3k sentences pieces, which is 0.6 F1-score lower than the \LA~ transition system. 
We added  examples in the appendix to illustrate mistakes in a Zero-shot setting (Figure~\ref{fig:zero-german}) and supervised English example (Figure~\ref{fig:english-example}). 

\subsection{Error Analysis}
\label{sec:error-analysis}

We observe two problems originating from the sequence-to-sequence models: first, hallucinations (words not found in the input) and second, ambiguous matches of mentions to the input. In order to evaluate the frequency of hallucinations, we counted cases where the predicted mentions and their context  could not be matched to a word sequence in the input. We found only 11 cases (0.07\%) in all 14.5k Link and Append predictions for the development set. 
The second problem are mentions with their n-gram context which are found more than once in the input. This cases constitutes 84 cases (0.6\%) of all 14.5k Link and Append predictions.

\begin{table}[h]
    \centering
    \small
    \begin{tabular}{l|r|r|r|r }
        Subset   & \#Docs & JS-L & CM &LA \\ \hline
        ~~~~1 -- 128  &  57 & 84.6 & 84.5 & 85.8 \\
        129 -- 256 & 73 & 83.7 & 83.6 & 85.2 \\
        257 -- 512 & 78 & 82.9 & 83.4 & 86.0 \\
        513 -- 768 & 71 & 80.1 & 79.3 & 83.2 \\
        769 -- 1152 & 52 & 79.1 & 78.6 & 83.3 \\
        1153+   & 12 & 71.3 & 69.6 & 74.9 \\ \hline
        all      & 343  & 80.1 &    79.5  &83.2 \\ 
         
    \end{tabular}
    \caption{Average F1-score on the development set for buckets of document length incremented by 128 tokens. The column JS-L shows average F1-scores for the SpanBert-Large model \cite{joshi-etal-2020-spanbert}, CM for the Constant Memory model \cite{xia-etal-2020-incremental} and LA for the Link-Append system.  The entries for JS-L and CM are taken from the paper of \newcite{xia-etal-2020-incremental}. 
    \label{tab:doc-length}}
    \label{tab:my_label}
\end{table}

Table \ref{tab:doc-length} shows average F1-scores
 for buckets of documents within a length range  incremented by 128 tokens, analogous to the analysis of \newcite{xia-etal-2020-incremental}. All system's F1-scores drop after the segment length 257--512 substantially by about 3-4 points. The Link-Append (LA) system  seems to have two more stable F1-score regions 1--512 and 513--1152 tokens divided by the mentioned larger drop while we see for the other system slightly lower accuracy in each segment.   

\subsection{Design Choice}

With this paper, we follow the paradigm of a text-2-text approach. Our goal was to use only the text output from a seq2seq model, and potentially the score associated with the output. Crucial for the high accuracy of the Link-Append systems are the design choices that seem to fit a text-2-text approach well.
(1) Initial experiments, not presented in the paper, showed lower performance for a standard two stage approach using mention prediction followed by mention-linking. The Link-only transition system which we included as a baseline in the paper was the first  system that we implemented that only predicted conference links, avoiding mention-detection. Hence this crucial first design choice is the  prediction of links and not to predict mentions first. (2) The prediction of links in a state-full fashion, where the prior input records previous coreference decisions, finally leads to the superior accuracy for the text-2-text model. 
(3) The larger model enables us to use the simpler paradigm of a text-2-text model successfully. The smaller models provide  substantially lower performance. We speculate in line with the arguments of \newcite{https://doi.org/10.48550/arxiv.2001.08361} that distinct capabilities of a model get strong or even emerge with  model size. 
(4) The strong multilingual results originates from the multilingual T5 model, which was initially surprising to us. For English, the mT5 model performed better as well which we attribute to larger vocab of the sentence piece encoding model of mT5.

\section{Conclusions}

In this paper, we combine a text-to-text (seq2seq) language model with a transition-based systems to perform coreference resolution. We reach 83.3 F1-score on the English CoNLL-2012 data set surpassing previous SotA. 
In the text-to-text framework, the \LA~transition system has been superior to hybrid \M~ transition system with mixed prediction of mentions, links and clusters. 
Our trained models are useful for future work as they could be used to initialize models for continuous training  or Zero-shot transfer to new languages. 

\section{Acknowledgments}

We would like to thank the action editor and three
anonymous reviewers for their thoughtful and
insightful comments, which were very helpful
in improving the paper.

\bibliography{tacl2021}
\bibliographystyle{acl_natbib}

\begin{figure*}[ht!]
\small
[1 Hausbesetzer " Prinz \textcolor{red}{\bf "} ] Hamburg ( ap ) - [1 Ein zwei Jahre alter Schäferhund namens " Prinz \textcolor{red}{\bf "} ] hat im Hamburger Stadtteil Altona [2 eine Wohnung ] besetzt . [3 Der 24jährige Besitzer ] hatte [1 dem Tier ] am Vortag [2  \textcolor{red}{\bf [3 sein ]} zukünftiges Heim ] \textcolor{red}{\bf [4 gezeigt ]}. \textcolor{red}{\bf [4 } \textcolor{red}{\bf Das} \textcolor{red}{\bf ]} gefiel [1 dem Hund ] so gut , daß [1 er ] unmittelbar hinter der Tür Stellung bezog und niemanden mehr durchließ. Als [5 ein Bekannter [3 des Hundehalters ] ] versuchte, \textcolor{red}{\bf die Wohnung} zu räumen, wurde [5 er ] gebissen und flüchtete ins Wohnzimmer zur Gattin [3 des Besitzers ] . Erst die Feuerwehr konnte beide durch das Fenster befreien. [3 Herrchen ] wollte den Hundefänger holen. 
\\
\caption{German Zero-shot predictions. The \textcolor{red}{\bf red bold} marked text are wrong predictions. 
\label{fig:zero-german}
}

~\\

In the summer of 2005 , a picture that people have long been looking forward to started emerging with frequency in various major [1 Hong Kong ] media . With [2 their ] unique charm , [2 these well - known cartoon images ] once again caused [1 Hong Kong ] to be a focus of worldwide attention . [4 The world 's fifth [3 Disney ] park ] will soon open to the public here . The most important thing about [3 Disney ] is that [3 it ] is a global brand . Well , for several years , although [4 it ] was still under construction and , er , not yet open , it can be said that many people have viewed [1 Hong Kong ] with new respect . Then welcome to the official writing ceremony of [4 \textcolor{red}{\bf Hong Kong} Disneyland ] . The construction of [4 \textcolor{red}{\bf Hong Kong} Disneyland ] began two years ago , in [5 2003 ] . In January of [5 that year ] , [1 \textcolor{red}{\bf the} Hong Kong \textcolor{red}{\bf government} ] turned over to [3 Disney Corporation ] [6 200 hectares of land at the foot of [7 Lantau Island ] that was obtained following the largest land reclamation project in recent years ] . One . Since then , [6 this area ] has become a prohibited zone in [1 Hong Kong ] . As [8 its ] neighbor on [7 Lantau Island ] , [8 \textcolor{red}{\bf Hong Kong} International Airport ] had to change [8 its ] flight routes to make [6 this area ] a no - fly zone . [4 Mickey Mouse 's new home ] , settling on Chinese land for the first time , has captured worldwide attention . There 's only one month left before the opening of [4 Hong Kong Disneyland ] on September 12 . The subway to [4 Disney ] has already been constructed . At subway stations , passengers will frequently press the station for [4 Disney ] on ticket machines , trying to purchase tickets to enjoy [4 the park ] when [4 it ] first opens .  Meanwhile , the \textcolor{red}{\bf [3 Disney ]} subway station is scheduled to open on the same day as [4 the park ] ...

\caption{Mistakes picked from CoNLL-2012 development set, e.g. {\em Hong Kong} should have been identified recursively within {\em Hong Kong Disneyland}; in the last sentence, {\em [3 Disney]} refers to {\em [3 Disney Corporation]} cluster instead correctly to {\em [4 The  world ’s fifth Disney park]} cluster. 
\label{fig:english-example}
}

~\\

... but it is pretty clear that there 's \textcolor{red}{\bf [18 a lot of blood ]} there \textcolor{red}{\bf [18 great deal of blood ]} . Larry Kobilinsky: \textcolor{red}{\bf [18 It ]} makes you think that something criminal occured . Larry Kobilinsky: But again getting back to how you solve [9 the crime ] you got to go to [12 the crime scene ] . Larry Kobilinsky: and [15 I ] think there are multiple scenes here . Larry Kobilinsky: Every place there 's blood it 's a crime scene . Larry Kobilinsky: But the most important place is [1 the cabin ] because [1 that ] 's presumably where the injury took place . Larry Kobilinsky: Now there was some rumor about some arguments going [21 casino ] the night before . Larry Kobilinsky: Uh and it could very well be that [20 he ] was arguing with [19 these individuals ] . Larry Kobilinsky: so uh that could be the tie the connection . Larry Kobilinsky: and [15 I ] think good policework would \textcolor{red}{\bf [22 connect ]} [19 the individuals in question ] with [21 the casino ] with [20 Mister Smith ] . Larry Kobilinsky: and \textcolor{red}{\bf [22 that ]} might help us understand what happened . Dan Abrams: [23 this ] is um a little bit more sound . Dan Abrams: [23 This ] is again from [3 Cletus Hyman uh who was literally um in a joining or nearby cabin \textcolor{red}{\bf um as to what} [3 he ] \textcolor{red}{\bf heard around} [3 he ] \textcolor{red}{\bf said four in the morning the day before uh } [20 Mister Smith ] \textcolor{red}{\bf went missing ]} . Cletus Hyman: At times it sounded like furniture was being actually picked up and dropped . Cletus Hyman: and then [24 that horrific uh thud ] . Dan Abrams: Yeah [7 I ] mean that sure sounds to [7 me ] [25 chief ] like we 're not talking about someone who /- Dan Abrams: [7 I ] mean [7 I ] guess it 's possible if you 're talking about loud yelling coming /- Dan Abrams: Well look if [3 this guy ] 's right that loud yelling is coming from [1 the cabin ] right and the wife is not a real suspect here means that there was probably someone else in there . ... Walter Zalisko: But it was five days later that \textcolor{red}{\bf [27 [13 the ship 's ] attorneys ]} had begun to question [26 her ] not law enforcement but \textcolor{red}{\bf [27 [13 ship 's ] attorneys ]} . Dan Abrams: We will uh continue to follow this . Dan Abrams: [25 Chief Zalisko of the Oakhill Police Department ] sorry about that [11 Susan Filan ] [15 Larry Kobilinsky ] thanks a lot .

\caption{Mistakes picked from CoNLL-2012 development, e.g., the coreferences {\em [18 a lot of blood]} as well as [27 [13 the ship ’s ] attorneys] are not in the gold annotation.  
\label{fig:english-example2}
}

\end{figure*}

\iftaclpubformat

\onecolumn

\appendix






  

\end{document}